\documentclass[conference]{IEEEtran}
\IEEEoverridecommandlockouts

\usepackage{cite}
\usepackage{graphicx}
\usepackage{amsmath,amssymb,amsfonts}
\usepackage{algorithmic}
\usepackage{textcomp}
\usepackage{xcolor}
\usepackage{amsthm}
\usepackage{booktabs}
\usepackage{hyperref}

\newtheorem{lemma}{Lemma}

\def\BibTeX{{\rm B\kern-.05em{\sc i\kern-.025em b}\kern-.08em
    T\kern-.1667em\lower.7ex\hbox{E}\kern-.125emX}}

\begin{document}

\title{SinSim: Sinkhorn-Regularized SimCLR\\
}

\author{\IEEEauthorblockN{M.~Hadi~Sepanj}
\IEEEauthorblockA{\textit{Vision Group, Systems Design Engineering} \\
\textit{University of Waterloo}\\
Waterloo, Canada \\
mhsepanj@uwaterloo.ca}
\and
\IEEEauthorblockN{Paul~Fieguth}
\IEEEauthorblockA{\textit{Vision Group, Systems Design Engineering} \\
\textit{University of Waterloo}\\
Waterloo, Canada \\
paul.fieguth@uwaterloo.ca}
}

\maketitle

\begin{abstract}
Self-supervised learning has revolutionized representation learning by eliminating the need for labeled data. Contrastive learning methods, such as SimCLR, maximize the agreement between augmented views of an image but lack explicit regularization to enforce a globally structured latent space. This limitation often leads to suboptimal generalization. We propose \textbf{SinSim}, a novel extension of SimCLR that integrates \textit{Sinkhorn regularization} from optimal transport theory to enhance representation structure. The Sinkhorn loss, an entropy-regularized Wasserstein distance, encourages a well-dispersed and geometry-aware feature space, preserving discriminative power. Empirical evaluations on various datasets demonstrate that SinSim outperforms SimCLR and achieves competitive performance against prominent self-supervised methods such as VICReg and Barlow Twins. UMAP visualizations further reveal improved class separability and structured feature distributions. These results indicate that integrating optimal transport regularization into contrastive learning provides a principled and effective mechanism for learning robust, well-structured representations. Our findings open new directions for applying transport-based constraints in self-supervised learning frameworks.
\end{abstract}


\section{Introduction}

Self-supervised representation learning has emerged as a powerful paradigm to learn effective visual features without the burden of manual annotations. Recent approaches such as SimCLR~\cite{chen2020simple}, BYOL~\cite{grill2020bootstrap}, VICReg~\cite{bardes2021vicreg}, and Barlow Twins~\cite{zbontar2021barlow} have demonstrated significant success by exploiting contrastive learning and redundancy reduction objectives. These methods have reshaped the landscape of representation learning, enabling models to match or even surpass supervised learning performance in certain tasks.

Despite their success, contrastive learning methods still face inherent limitations. SimCLR, for instance, primarily relies on pairwise similarity maximization, which ensures that different augmentations of the same image remain close in the latent space. However, this approach does not necessarily promote a globally well-structured representation space, which can lead to poor generalization and mode collapse \cite{ziyin2022loss}. Without an explicit regularization mechanism, representations may cluster too tightly or fail to maintain meaningful global relationships among different samples.

To address this, we propose \textbf{SinSim}, a novel extension of SimCLR~\cite{chen2020simple} that integrates \textbf{Sinkhorn regularization} ~\cite{wang2021sinkhorn,cuturi2013sinkhorn} from optimal transport theory into the contrastive learning framework. Optimal transport provides a principled way to align distributions while preserving global structure, making it a natural choice for contrastive learning. The Sinkhorn loss, an entropy-regularized version of the Wasserstein distance \cite{villani2009optimal,kantorovich1942transfer}, enforces a structured and well-dispersed latent space by promoting geometric consistency. This results in feature representations that are not only more discriminative but also more robust to variations in data.

The key contributions of this paper are:
\begin{itemize}
    \item We introduce \textbf{Sinkhorn regularization} into the SimCLR framework to enhance the global structure of learned representations.
    \item We provide a \textbf{theoretical justification}, showing how the Sinkhorn loss acts as a geometry-aware regularizer.
    \item We conduct experiments on standard benchmarks, demonstrating that SinSim outperforms SimCLR in both linear and non-linear classification tasks.
\end{itemize}

The following sections describe our methodology, present theoretical insights, and validate our approach through empirical results.

\section{Background}

\subsection{Self-Supervised Contrastive Learning}
Self-supervised learning has emerged as a dominant paradigm for learning representations without the need for manual annotations. Among various self-supervised learning methods, \textit{contrastive learning} has proven to be particularly effective in visual representation learning~\cite{chen2020simple, sepanj2023context, grill2020bootstrap}. The core idea behind contrastive learning is to bring similar instances (positive pairs) closer in the latent space while pushing apart dissimilar instances (negative pairs). 

One of the most well-known frameworks, SimCLR~\cite{chen2020simple}, achieves contrastive learning by maximizing the agreement between different augmentations of the same image. Formally, given a mini-batch of $N$ images, two augmentations of each image are generated, resulting in $2N$ samples. The representations of these samples are mapped to a latent space and trained with the normalized temperature-scaled cross-entropy loss (NT-Xent) \cite{chen2020simple}. This loss encourages representations of the same image (positive pairs) to be aligned while ensuring diversity by repelling different images (negative pairs). However, SimCLR heavily relies on instance discrimination, where each image is treated as its own class, ensuring that different augmentations of the same image are pulled together while all other images are pushed apart. This can lead to suboptimal global structure, as it does not explicitly encourage relationships between semantically similar images.

While methods such as MoCo~\cite{he2020momentum} and BYOL~\cite{grill2020bootstrap} have attempted to mitigate these issues by employing momentum encoders and redundancy reduction, the lack of an explicit global regularizer remains a limitation. We argue that incorporating an optimal transport-based regularization term can explicitly structure the representation space while preserving discriminative power.

\subsection{Sinkhorn Regularization and Optimal Transport}

Optimal transport (OT) provides a principled framework for comparing probability distributions in a geometrically meaningful way~\cite{alvarez2020geometric,cuturi2013sinkhorn,villani2009optimal}. Unlike conventional similarity measures, OT considers the minimal \textit{transport cost} required to morph one distribution into another. The classic Wasserstein distance \cite{villani2009optimal,kantorovich1942transfer}, also known as the Earth Mover’s Distance (EMD), quantifies this cost as
\begin{equation}
    W_c(\mu, \nu) = \inf_{\gamma \in \Pi(\mu, \nu)} \sum_{i,j} c(x_i, y_j) \gamma_{ij},
\end{equation}
where $\Pi(\mu, \nu)$ represents the set of joint probability distributions (or \emph{couplings}) with marginals $\mu$ and $\nu$, $c(x_i, y_j)$ defines the cost to transport mass between samples, and $\gamma$ is a transport plan whose entry $\gamma_{ij}$ specifies the amount of mass transported from $x_i$ to $y_j$.

Although powerful, the direct computation of the Wasserstein distance is computationally prohibitive due to its reliance on solving a linear program. To address this, \textbf{Sinkhorn regularization} \cite{wang2021sinkhorn,cuturi2013sinkhorn} introduces an entropy term that smooths the transport plan and enables efficient optimization:
\begin{equation}
    W_\lambda(\mu, \nu) = \inf_{\gamma \in \Pi(\mu, \nu)} \sum_{i,j} c(x_i, y_j) \gamma_{ij} + \lambda H(\gamma),
\end{equation}
where $H(\gamma) = -\sum_{i,j} \gamma_{ij} \log \gamma_{ij}$ is the entropy term and $\lambda > 0$ controls the trade-off between the transport cost and the degree of smoothing. This entropy regularization not only improves numerical stability but also leads to a differentiable approximation of the transport distance. The \textit{Sinkhorn-Knopp algorithm}~\cite{cuturi2013sinkhorn} iteratively normalizes the rows and columns of the transport plan to meet the marginal constraints, and it converges rapidly in practice, making it particularly suitable for deep learning applications.

\subsection{Related Work}
The integration of optimal transport into representation learning has gained traction in recent years. Several works have explored the use of Wasserstein distance for deep learning tasks such as domain adaptation~\cite{courty2017optimal, flamary2016optimal, damodaran2018deepjdot}, generative modeling~\cite{arjovsky2017wasserstein,tolstikhin2018wasserstein,gulrajani2017improved}, and clustering~\cite{genevay2019learning,fatras2021unbalanced,cuturi2014fast}. In self-supervised learning, optimal transport has been investigated as a tool for feature alignment~\cite{fatras2021unbalanced,perrot2016mapping,schmitz2018wasserstein} and structured representation learning~\cite{patrini2020sinkhorn,genevay2018learning,asano2019self}.

Contrastive learning methods traditionally rely on predefined similarity measures such as cosine similarity or Euclidean distance~\cite{kumar2022contrastive}. However, these metrics do not capture the global structure of learned representations. Recent efforts such as CDSSL~\cite{sepanj2025self}, MMD-Based VICReg~\cite{sepanj2024aligning} ,VICReg~\cite{bardes2021vicreg} and Barlow Twins~\cite{zbontar2021barlow} introduced alternative regularization strategies based on variance-covariance matrices to enhance representation dispersion. Our approach builds upon these insights by leveraging Sinkhorn-based optimal transport to enforce a structured latent space while maintaining the contrastive learning framework of SimCLR.

In contrast to prior work, our proposed method, \textbf{SinSim}, explicitly integrates Sinkhorn loss into SimCLR, bridging the gap between instance-based alignment and global structure regularization. We hypothesize that this hybrid approach not only prevents embedding collapse but also improves the overall robustness of learned representations.

\section{Methodology}

In this section, we detail the architecture and training procedure of our proposed \textbf{SinSim} framework, emphasizing the distinct roles of the intermediate representation \( h \) and the final embedding \( z \).

\subsection{Network Architecture}

Our model comprises two primary components:
\begin{enumerate}
\item The \textbf{Encoder Network \( f_{\theta} \)} transforms the input data \( x \) into intermediate representations \( h \):
   \[
   h = f_{\theta}(x)
   \]
   These representations capture essential features of the input data.

\item The \textbf{Projection Head \( g_{\phi} \)} maps \( h \) to the final embedding space \( z \):
   \[
   z = g_{\phi}(h)
   \]
   The projection head is designed to facilitate the learning of embeddings suitable for contrastive objectives.
\end{enumerate}

\subsection{Training Objective}

The training process involves two key components.  First, there is the  \textbf{Contrastive Loss on \( z \)}, in which we apply a contrastive loss, such as the NT-Xent loss \cite{chen2020simple}, on the embeddings \( z_1 \) and \( z_2 \) obtained from different augmentations of the same input. This loss encourages the model to bring positive pairs closer and push negative pairs apart in the embedding space:
\begin{equation}
\mathcal{L}_{\text{contrastive}} = -\log \frac{\exp(\text{sim}(z_1, z_2)/\tau)}{\sum_{k=1}^{2N} \mathbb{1}_{[k \neq i]} \exp(\text{sim}(z_i, z_k)/\tau)}
\label{eqn-xent}
\end{equation}
where \( \text{sim}(\cdot, \cdot) \) denotes cosine similarity, \( \tau \) is a temperature parameter, and \( N \) is the batch size.

Next, to enforce a well-structured representation space and prevent collapse, we introduce a \textbf{Sinkhorn Regularization} on the intermediate representations \( h_1 \) and \( h_2 \) corresponding to the embeddings \( z_1 \) and \( z_2 \). The Sinkhorn loss \cite{cuturi2013sinkhorn} is defined as
\begin{equation}
\mathcal{L}_{\text{Sinkhorn}} = \min_{\gamma \in \Pi(\mu, \nu)} \sum_{i,j} \gamma_{ij} c(h_1^{(i)}, h_2^{(j)}) + \lambda H(\gamma)   
\label{eqn-sink}
\end{equation}
where \( \Pi(\mu, \nu) \) denotes the set of joint probability distributions with marginals \( \mu \) and \( \nu \), \( c \) is a cost function (e.g., squared Euclidean distance), \( \lambda \) is a regularization parameter, and \( H(\gamma) \) is the entropy of the transport plan \( \gamma \).
The total loss is then a weighted combination of these two terms:
\begin{equation}\label{eq_loss}
\mathcal{L} = \mathcal{L}_{\text{contrastive}} + \beta \mathcal{L}_{\text{Sinkhorn}}    
\end{equation}
where \( \beta \) controls the influence of the Sinkhorn regularization.

\subsection{Rationale for Regularization on \( h \) and Clustering on \( z \)}

Applying Sinkhorn regularization on the intermediate representations \( h \) rather than the final embeddings \( z \) offers several advantages:
\begin{itemize}
\item \textbf{Preservation of Information:} The representations \( h \) retain more information about the input data, as they are directly produced by the encoder \( f_{\theta} \). Regularizing \( h \) ensures that the encoder learns to distribute information uniformly across the representation space, promoting diversity and preventing collapse.

\item \textbf{Flexibility in Embedding Space:} The projection head \( g_{\phi} \) is tasked with mapping \( h \) to \( z \) in a way that is conducive to the contrastive objective. By applying clustering objectives on \( z \), we allow \( g_{\phi} \) to focus on organizing the embeddings for better class separation without being constrained by the regularization applied to \( h \).

\item \textbf{Computational Efficiency:} The dimensionality of \( h \) is typically lower than that of \( z \), making the computation of the Sinkhorn loss more efficient when applied to \( h \).
\end{itemize}
This overall design strategy, to regularize the intermediate representation, aligns with recent findings in contrastive learning literature \cite{bardes2021vicreg}, where decoupling the roles of different network components has been shown to enhance performance.

\subsection{Implementation Details}

Our training framework follows the SimCLR paradigm, where each input sample undergoes two distinct augmentations to create positive pairs. These augmented views are then processed through a {ResNet-18 encoder}, followed by a {projection head}, producing intermediate representations \( h_1, h_2 \) and projected embeddings \( z_1, z_2 \).

The \textbf{NT-Xent loss} from (\ref{eqn-xent}) ensures contrastive alignment in the final embedding space, while the \textbf{Sinkhorn loss} (\ref{eqn-sink}) enforces a structured and well-dispersed representation space at the intermediate feature level. 

During training, we initialize the model with random weights and update parameters using the {Adam optimizer} with a learning rate of \( 10^{-3} \) and weight decay of \( 10^{-6} \). The encoder is trained for a total of 100 epochs on the MNIST dataset and 300 epochs for the rest of the datasets using a batch size of 512. Following pretraining, the encoder's weights are frozen, and a linear classifier is trained on top of the learned representations for downstream evaluation.

The hyperparameters \( \beta \) and \( \lambda \) are tuned based on validation performance. The Sinkhorn distance is computed iteratively using 40 updates with the selected \( \lambda = 0.05 \) from validation to stabilize the optimal transport plan.

In practice, we compute the Sinkhorn loss using the \emph{Sinkhorn--Knopp algorithm}~\cite{cuturi2013sinkhorn}, as detailed below:
\begin{enumerate}
    \item \textbf{Cost Matrix Computation:} For a mini-batch of representations, compute the cost matrix
    \[
    C_{ij} = \|z_1^{(i)} - z_2^{(j)}\|^2.
    \]
    \item \textbf{Initialization:} Initialize the coupling \(\gamma\) to an isotropic matrix (e.g., all entries equal to \(1/N^2\)).
    \item \textbf{Iterative Scaling:} Update the coupling iteratively via
    \[
    \gamma^{(t+1)} = \operatorname{diag}(u^{(t)})\,\gamma^{(t)}\,\operatorname{diag}(v^{(t)}),
    \]
    where the scaling vectors \(u^{(t)}\) and \(v^{(t)}\) are computed to enforce the row and column marginal constraints.
    \item \textbf{Loss Computation:} Upon convergence, the stabilized transport plan is used to compute the final Sinkhorn loss.
\end{enumerate}


\subsection{Theoretical Justification}

It is important to  provide a justification for the use of Sinkhorn regularization in our framework. Recall that for two empirical distributions
\[
P = \frac{1}{N}\sum_{i=1}^{N}\delta_{z_1^{(i)}}, \quad Q = \frac{1}{N}\sum_{i=1}^{N}\delta_{z_2^{(i)}},
\]
and a cost matrix with entries
\[
C_{ij} = \|z_1^{(i)} - z_2^{(j)}\|^2,
\]
the entropy-regularized Wasserstein distance is defined as
\begin{equation}
W_\lambda(P,Q) = \min_{\gamma \in \Pi(P,Q)} \; \langle \gamma, C \rangle - \lambda H(\gamma),
\label{eq:W_lambda}
\end{equation}
where
\[
H(\gamma) = -\sum_{i,j} \gamma_{ij}\log\gamma_{ij},
\]
and \(\Pi(P,Q)\) is the set of couplings with marginals \(P\) and \(Q\).

\begin{lemma}[Latent Space Dispersion]
Let \(P\) and \(Q\) be defined as above, and denote by \(\gamma^*\) the optimal coupling in \eqref{eq:W_lambda}. Then,
\begin{equation}
W_\lambda(P,Q) \geq \frac{1}{N} \sum_{i=1}^{N} \| z_1^{(i)} - z_2^{(i)} \|^2 - \lambda H(\gamma^*).
\label{eq:dispersion}
\end{equation}
In particular, as \(\lambda\to 0\), minimizing \(W_\lambda(P,Q)\) forces the learned representations to align (by the``diagonal'' cost) while still controlling dispersion in the latent space.
\end{lemma}

\begin{proof}
\textbf{1. Candidate Coupling:}  
Consider the coupling given by
\[
\tilde{\gamma}_{ij} = \frac{1}{N}\delta_{ij},
\]
where \(\delta_{ij}\) is the Kronecker delta. This coupling satisfies the marginal constraints in \(\Pi(P,Q)\).

\textbf{2. Evaluation of the Objective:}  
For \(\tilde{\gamma}\), the cost term is
\[
\langle \tilde{\gamma}, C \rangle = \frac{1}{N}\sum_{i=1}^N \|z_1^{(i)} - z_2^{(i)}\|^2.
\]
Its entropy is computed as
\[
H(\tilde{\gamma}) = -\sum_{i=1}^N \frac{1}{N} \log\left(\frac{1}{N}\right) = \log N.
\]
Thus, the regularized cost evaluated at \(\tilde{\gamma}\) is
\[
\langle \tilde{\gamma}, C \rangle - \lambda H(\tilde{\gamma}) = \frac{1}{N}\sum_{i=1}^N \|z_1^{(i)} - z_2^{(i)}\|^2 - \lambda \log N.
\]

\textbf{3. Lower Bound via Optimality:}  
Since \(\gamma^*\) minimizes the objective in \eqref{eq:W_lambda}, we have
\begin{align*}
W_\lambda(P,Q) = \langle \gamma^*, C \rangle &- \lambda H(\gamma^*) \\\leq 
&\frac{1}{N}\sum_{i=1}^N \|z_1^{(i)} - z_2^{(i)}\|^2 - \lambda \log N.
\end{align*}
However, note that for any \(\gamma \in \Pi(P,Q)\), the cost term satisfies
\[
\langle \gamma, C \rangle \geq \frac{1}{N}\sum_{i=1}^N \|z_1^{(i)} - z_2^{(i)}\|^2,
\]
since any deviation from the ``diagonal'' pairing increases the quadratic cost. In particular, for the optimal \(\gamma^*\) we obtain
\begin{align*}
W_\lambda(P,Q) &= \langle \gamma^*, C \rangle - \lambda H(\gamma^*) \\
&\geq \frac{1}{N}\sum_{i=1}^N \|z_1^{(i)} - z_2^{(i)}\|^2 -\lambda H(\gamma^*).
\end{align*}

This establishes \eqref{eq:dispersion}.

\textbf{4. Interpretation:}  
As \(\lambda\to 0\), the penalty term \(\lambda H(\gamma^*)\) becomes negligible, and the minimization of \(W_\lambda(P,Q)\) forces the cost \(\langle \gamma^*, C \rangle\) to approach the average diagonal cost. This structured alignment ensures dispersion in the latent space.
\end{proof}

\begin{lemma}[Prevention of Mode Collapse]
Assume that the contrastive loss \(\mathcal{L}_{\text{NT-Xent}}\) is minimized, and let \(\gamma^*\) denote the optimal transport plan obtained via the Sinkhorn optimization. Then, for any \(\lambda > 0\) (in particular, for sufficiently small \(\lambda\)), the strong convexity induced by the entropy term guarantees that
\begin{equation}
\gamma^*_{ij} > 0 \quad \forall\, i,j.
\label{eq:mode_collapse}
\end{equation}
\end{lemma}

\begin{proof}
\textbf{1. Optimality Conditions:}  
The inclusion of the entropy term in \eqref{eq:W_lambda} makes the optimization problem strictly convex. Consequently, the optimal coupling \(\gamma^*\) is unique and is characterized by the Sinkhorn fixed-point equations.

\textbf{2. Gibbs Form Representation:}  
It is well established (see, e.g., \cite{cuturi2013sinkhorn}) that the solution \(\gamma^*\) can be written in Gibbs form as
\begin{equation}
\gamma^*_{ij} = \exp\!\left(\frac{f_i + g_j - C_{ij}}{\lambda}\right),
\label{eqn-gibbs}
\end{equation}
where \(f_i\) and \(g_j\) are dual potentials corresponding to the marginal constraints. 

\textbf{3. Strict Positivity:}  
Since the exponential function is strictly positive for any finite argument, from (\ref{eqn-gibbs}) it follows that
\[
\gamma^*_{ij} > 0 \quad \forall\, i,j.
\]
Thus, the entropic regularization prevents any entry of the transport plan from collapsing to zero, thereby mitigating mode collapse.
\end{proof}

\section{Experiments}
\label{sec:experiments}

\begin{figure*}[h]
    \centering
    \begin{tabular}{cccc}
        \includegraphics[width=0.23\textwidth]{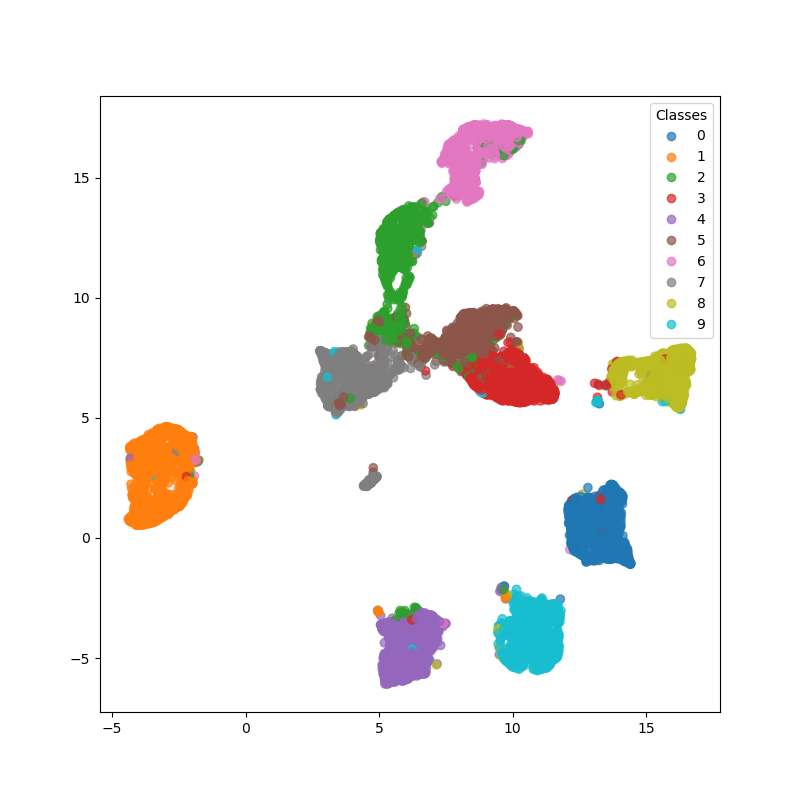} & 
        \includegraphics[width=0.23\textwidth]{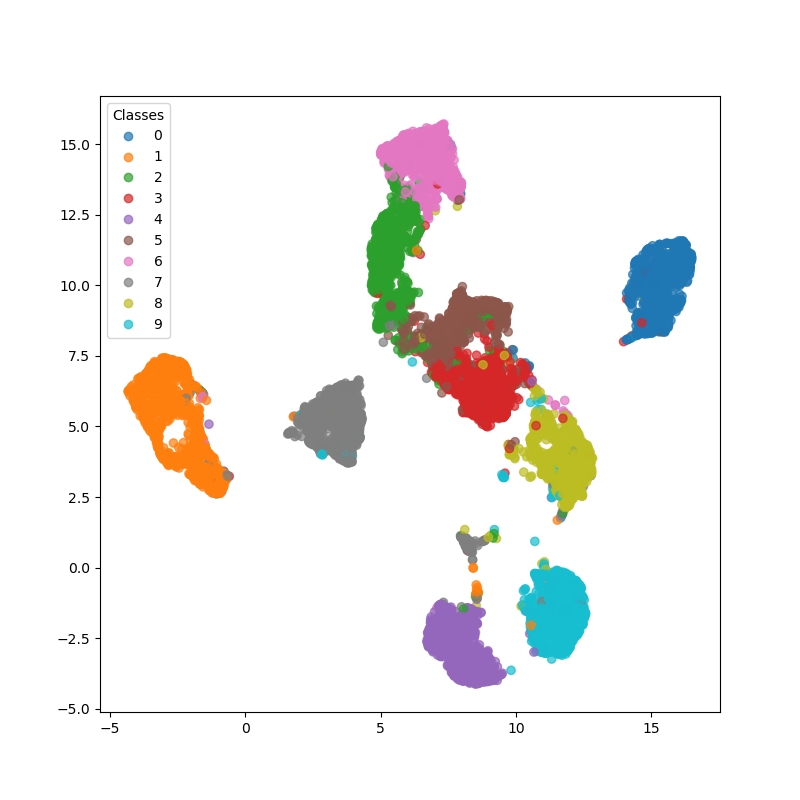} &
        \includegraphics[width=0.23\textwidth]{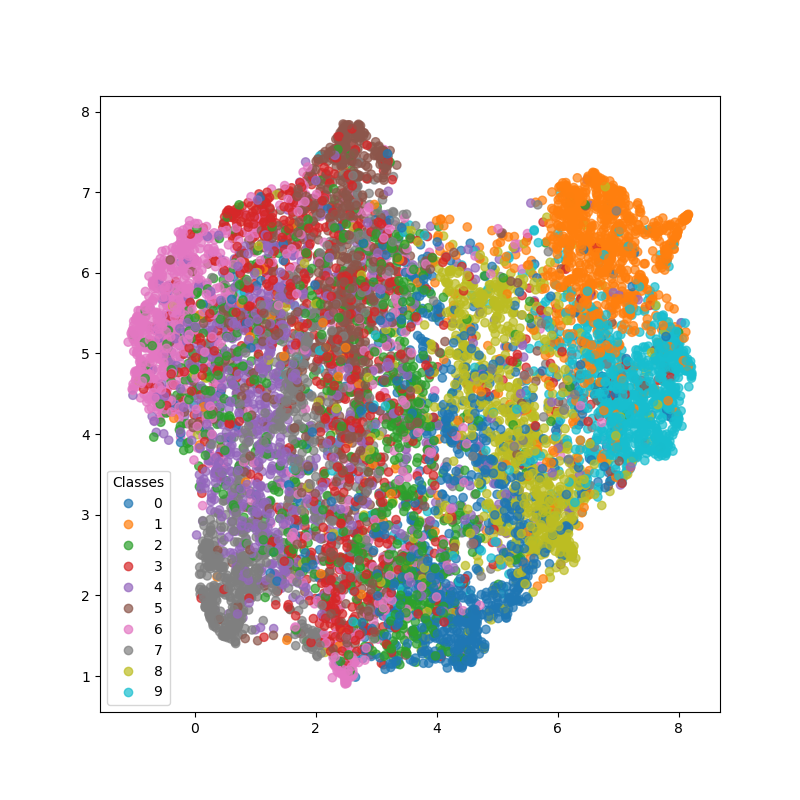} & 
        \includegraphics[width=0.23\textwidth]{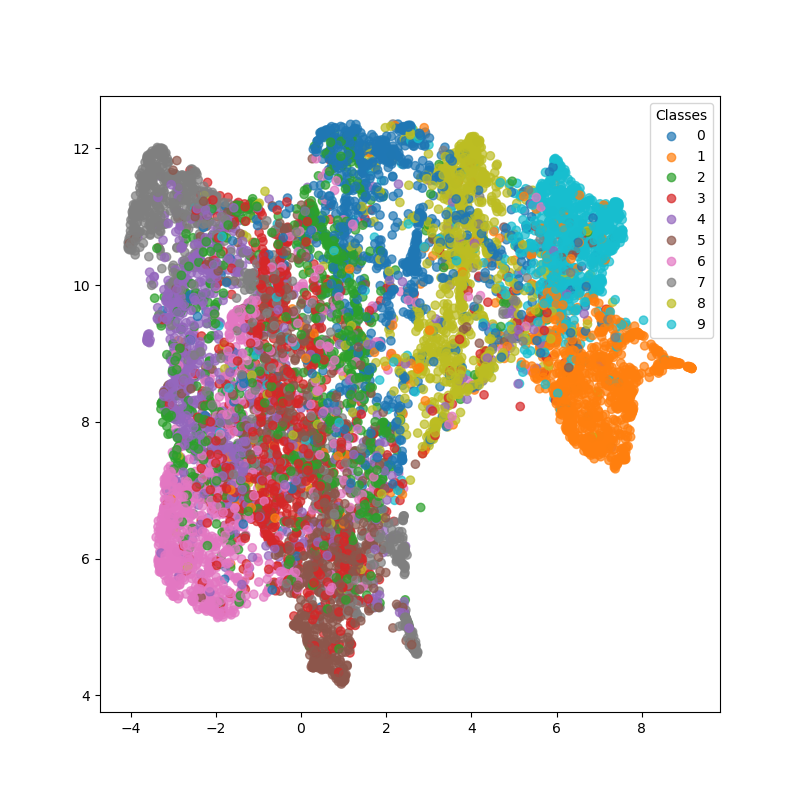} \\
        (a) SimCLR - MNIST & (b) SinSim - MNIST & (c) SimCLR - CIFAR-10 & (d) SinSim - CIFAR-10 \\
    \end{tabular}
    \caption{UMAP visualization of learned embeddings. (a) and (c) show the embedding space for SimCLR, while (b) and (d) illustrate SinSim's feature representations. SinSim achieves better class separation and reduced overlap, suggesting that Sinkhorn regularization improves the structured alignment of representations.}
    \label{fig:representation_visualization}
\end{figure*}

We evaluate \textbf{SinSim} on multiple benchmark datasets, including \textbf{MNIST, CIFAR-10, CIFAR-100, and STL-10}, to assess its effectiveness in self-supervised representation learning. Our approach is compared against \textbf{SimCLR}~\cite{chen2020simple}, \textbf{VICReg}~\cite{bardes2021vicreg}, and \textbf{Barlow Twins}~\cite{zbontar2021barlow}, which represent prominent contrastive and redundancy reduction methods. 

Unlike these baselines, \textbf{SinSim} integrates Sinkhorn regularization, enforcing structured alignment in the latent space while preserving discriminative power. Through our experiments, we will demonstrate that \textbf{SinSim} not only enhances representation quality but also achieves comparable performance in both \textit{linear and non-linear classification tasks}, benefiting particularly from the global structure imposed by optimal transport.

\subsection{Experimental Setup}
We perform experiments on the MNIST, CIFAR-10, CIFAR-100, and STL-10 datasets. All models use a ResNet-18 backbone with an additional projection head for the contrastive learning task. Following pretraining, the encoder is frozen and evaluated with both a linear classifier and a two-layer MLP.

\subsection{Linear Classification Performance}

Table~\ref{tab:linear_classification} summarizes the linear classification accuracies across different self-supervised learning methods. Our proposed approach, \textbf{SinSim}, outperforms SimCLR on {\em every} dataset, and demonstrates competitive performance against VICReg and Barlow Twins, in particular with the proposed SinSim achieving the highest classification accuracy on both MNIST and STL-10.

\begin{table}[h]
    \centering
    \caption{Linear Classification Accuracy (\%)}
    \label{tab:linear_classification}
    \begin{tabular}{lcccc}
        \toprule
        Method & MNIST & CIFAR-10 & CIFAR-100 & STL-10 \\
        \midrule
        SimCLR & 97.0 & 68.6 & 41.9 & 50.3 \\
        VICReg & 96.1 & \textbf{72.4} & 45.5 & 52.4 \\
        Barlow Twins & 95.4 & 72.1 & \textbf{46.1} & 51.3 \\
        \textbf{SinSim (Ours)} & \textbf{98.4} & 70.7 & 43.2 & \textbf{52.6} \\
        \bottomrule
    \end{tabular}
\end{table}

While SinSim does not surpass VICReg on CIFAR-10, it still performs comparably, showing that Sinkhorn regularization does not degrade contrastive learning performance and provides an alternative to variance-covariance-based regularization strategies.
 
The improvement of SinSim over SimCLR, particularly on MNIST (+1.4\%) and STL-10 (+2.3\%), highlights the benefits of enforcing a more structured representation space via optimal transport. Moreover, while VICReg and Barlow Twins rely on statistical regularization (variance-covariance constraints) to encourage representation spread, Sinkhorn regularization explicitly models the geometry-aware distribution alignment, making SinSim a meaningful alternative in contrastive learning frameworks.



\subsection{Non-Linear Classification Performance}
We further assess the quality of the learned representations with a non-linear (MLP) classifier, shown in Table~\ref{tab:nonlinear_classification}.  With results very much consistent of those in Table~\ref{tab:linear_classification},  Table~\ref{tab:nonlinear_classification} demonstrates that the benefits of the SinSim strategy remain present with nonlinear classification, with SinSim still improving upon the baseline methods, suggesting that the global structure imposed by the Sinkhorn loss aids in learning more semantically meaningful features.

\begin{table}[h]
    \centering
    \caption{Non-Linear Classification Accuracy (\%)}
    \label{tab:nonlinear_classification}
    \begin{tabular}{lcccc}
        \toprule
        Method & MNIST & CIFAR-10 & CIFAR-100 & STL-10 \\
        \midrule
        SimCLR & 97.7 & 70.8 & 42.1 & 52.6 \\
        VICReg &  97.2 & \textbf{74.5} & \textbf{46.2} & 53.9 \\
        Barlow Twins & 96.9 & 73.6 & 46.0 & 52.2 \\
        \textbf{SinSim (Ours)} & \textbf{98.9} & 72.6 & 43.8 & \textbf{54.4} \\
        \bottomrule
    \end{tabular}
\end{table}

\subsection{Representation Visualization}

To further investigate the effect of Sinkhorn regularization on learned representations, we apply UMAP \cite{maaten2008visualizing} to the embeddings extracted from different models. Figure~\ref{fig:representation_visualization} visualizes the latent space for MNIST and CIFAR-10, comparing SimCLR and SinSim.

Comparing the embeddings obtained using SimCLR (panels a, c) and SinSim (panels b, d), we observe:
\begin{itemize}
    \item \textbf{Improved Class Separation:} SinSim produces more distinct clusters in MNIST and CIFAR-10, demonstrating that Sinkhorn regularization prevents mode collapse and encourages well-distributed representations.
    \item \textbf{Reduced Class Overlap:} In CIFAR-10, SimCLR embeddings (panel c) show overlap between different classes, whereas SinSim (panel d) results in more compact and well-separated clusters.
\end{itemize}
These findings provide qualitative evidence that the Sinkhorn regularization enforces structured alignment in contrastive learning, leading to more robust, well-separated, and meaningful representations.



\subsection{Ablation Study}

\subsubsection*{Effect of Sinkhorn Regularization}

To analyze the impact of Sinkhorn regularization on the representation learning process in contrastive learning, we conduct an ablation study by varying the strength \( \beta \) of the Sinkhorn regularization.

\begin{figure}[t]
    \centering
    \includegraphics[width=0.5\textwidth]{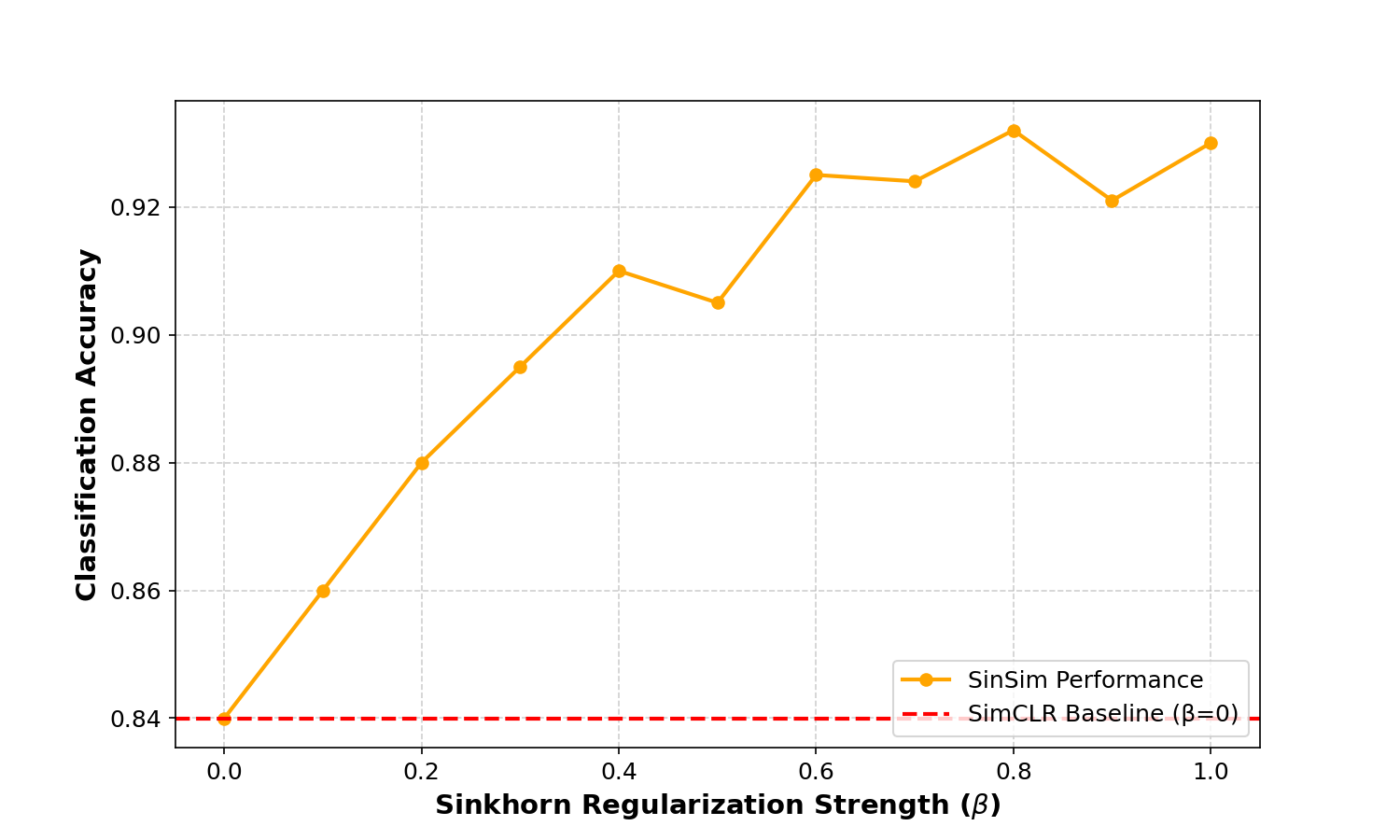}
    \caption{Effect of Sinkhorn Regularization on SimSim Performance, varying the Sinkhorn regularization strength (\(\beta\)) on classification accuracy when training on MNIST for \textbf{10 epochs}. The \textbf{solid orange curve} represents the classification accuracy achieved by the SinSim model as a function of \(\beta\), while the \textbf{dashed red line} corresponds to the baseline performance of standard SimCLR (equivalent to SinSim at \(\beta = 0\)). Very clearly, incorporating Sinkhorn regularization consistently improves feature representations, leading to enhanced classification accuracy. This preliminary experiment primarily serves to capture the trend of Sinkhorn's influence on contrastive learning rather than as an absolute performance benchmark.}
    \label{fig:sinkhorn_simclr}
\end{figure}

\begin{figure}[t]
    \centering
    \includegraphics[width=0.5\textwidth]{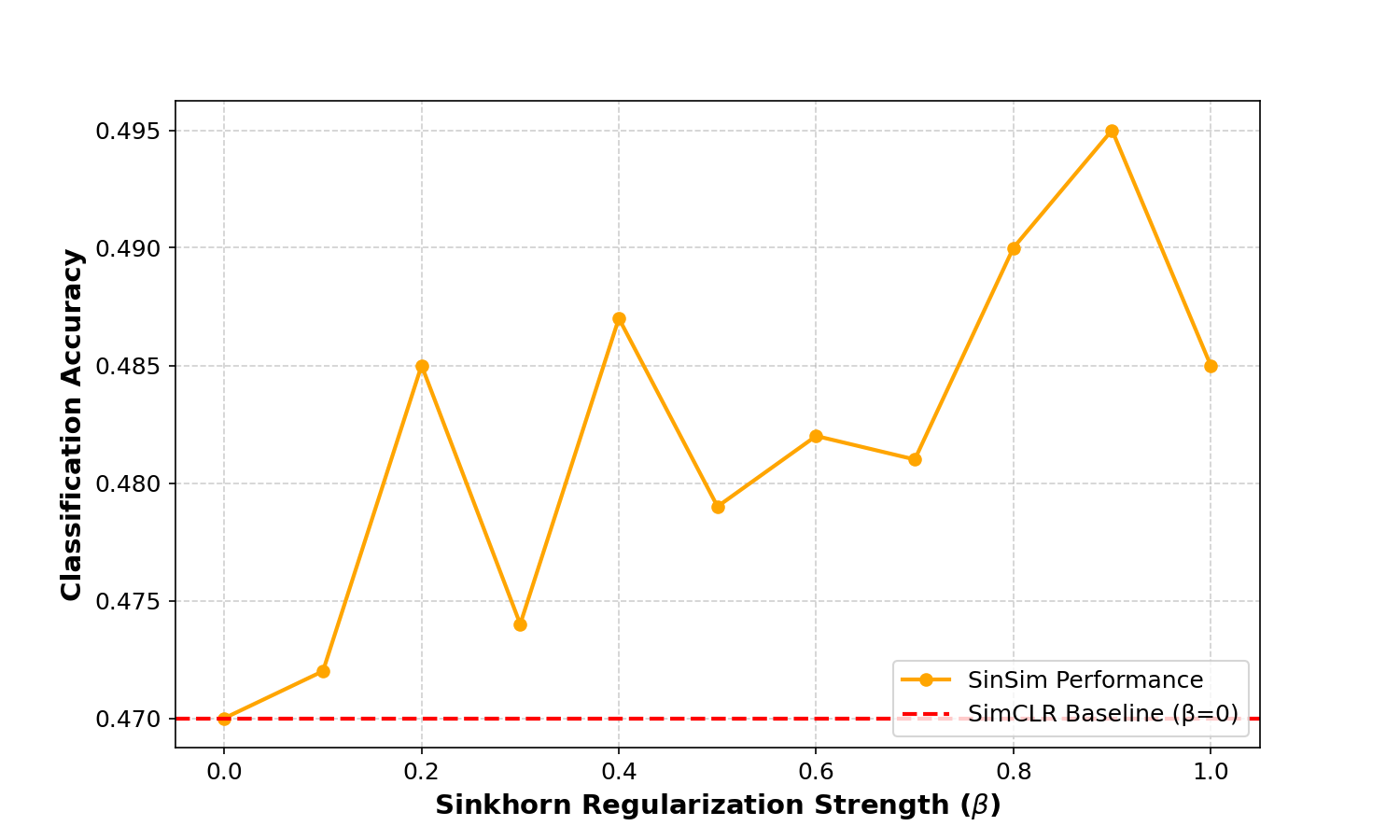}
    \caption{As in Figure~\ref{fig:sinkhorn_simclr}, but here based on CIFAR-10 data. Unlike the consistent improvement observed in Figure~\ref{fig:sinkhorn_simclr} with MNIST, here the results based on CIFAR-10 exhibit significant fluctuations with $\beta$, with a more modest trend. Nevertheless, for all values of $\beta$ SinSim outperforms the baseline, suggesting that Sinkhorn regularization definitely contributes to enhanced feature representations. This experiment primarily serves to capture the trend of Sinkhorn's influence on contrastive learning in a more complex dataset.}
    \label{fig:sinkhorn_simclr_2}
\end{figure}

Figures~\ref{fig:sinkhorn_simclr} and~\ref{fig:sinkhorn_simclr_2} present the classification accuracy of a linear probe trained on the representations learned by SinSim as a function of \( \beta \).  Both SinSim and standard SimCLR (effectively SinSim at \(\beta = 0\)) are pretrained on the MNIST and CIFAR-10 datasets for 10 epochs, followed by the training of a linear classifier for 5 epochs.

The crux of our analysis is that the  results unambigously demonstrate an improvement in classification accuracy as \( \beta \) increases, suggesting that the presence (larger $\beta$) of Sinkhorn regularization enhances the feature space structure by promoting better alignment of representations. The classification accuracy improves significantly for moderate values of \( \beta \), peaking around \( \beta = 0.8 \) for MNIST and \( \beta = 0.9 \) for CIFAR-10. This trend highlights the potential of optimal transport regularization in improving contrastive learning objectives by refining feature representation structure.

\subsubsection*{Effect of Sinkhorn Iterations}

The number of Sinkhorn iterations controls the convergence of the optimal transport plan. A small number of iterations may lead to an under-optimized transport plan, while too many iterations may introduce excessive regularization. 

\begin{figure}[t]
    \centering
    \includegraphics[width=1\linewidth]{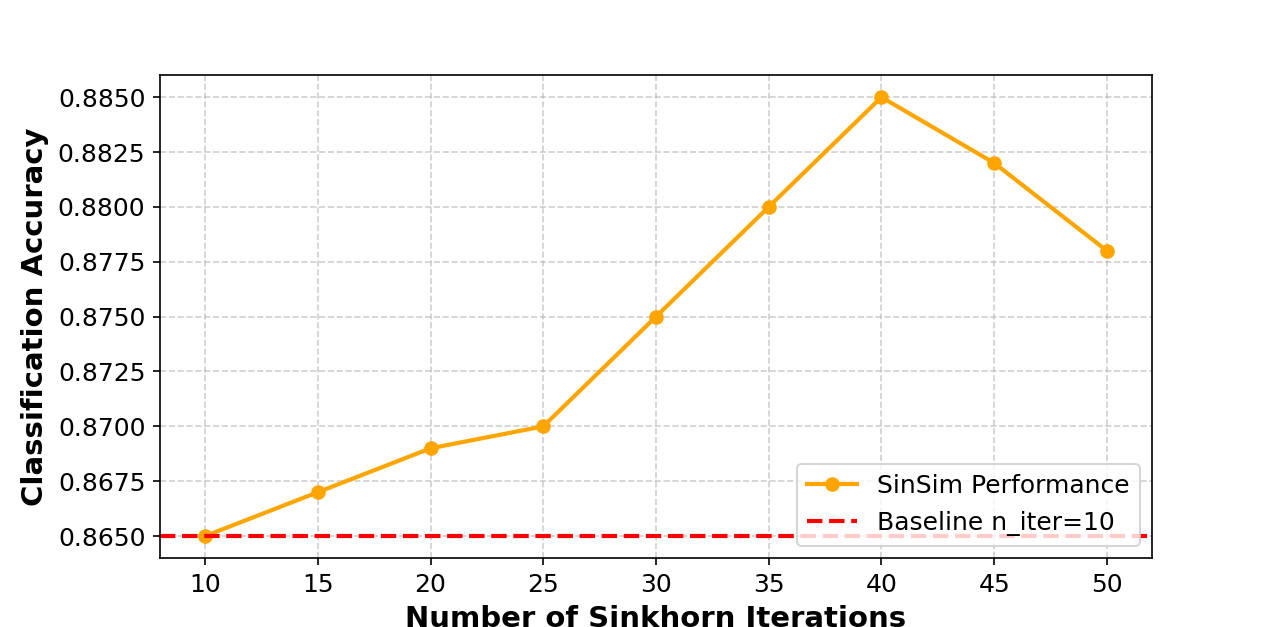}
    \caption{Effect of Sinkhorn iterations on SinSim classification accuracy on MNIST. Increasing the number of iterations improves accuracy up to 40, beyond which performance slightly decreases. The red dashed line represents a baseline SinSim performance at a default iteration count of 10.}
    \label{fig:n_iter}
\end{figure}

\begin{figure}[t]
    \centering
    \includegraphics[width=1\linewidth]{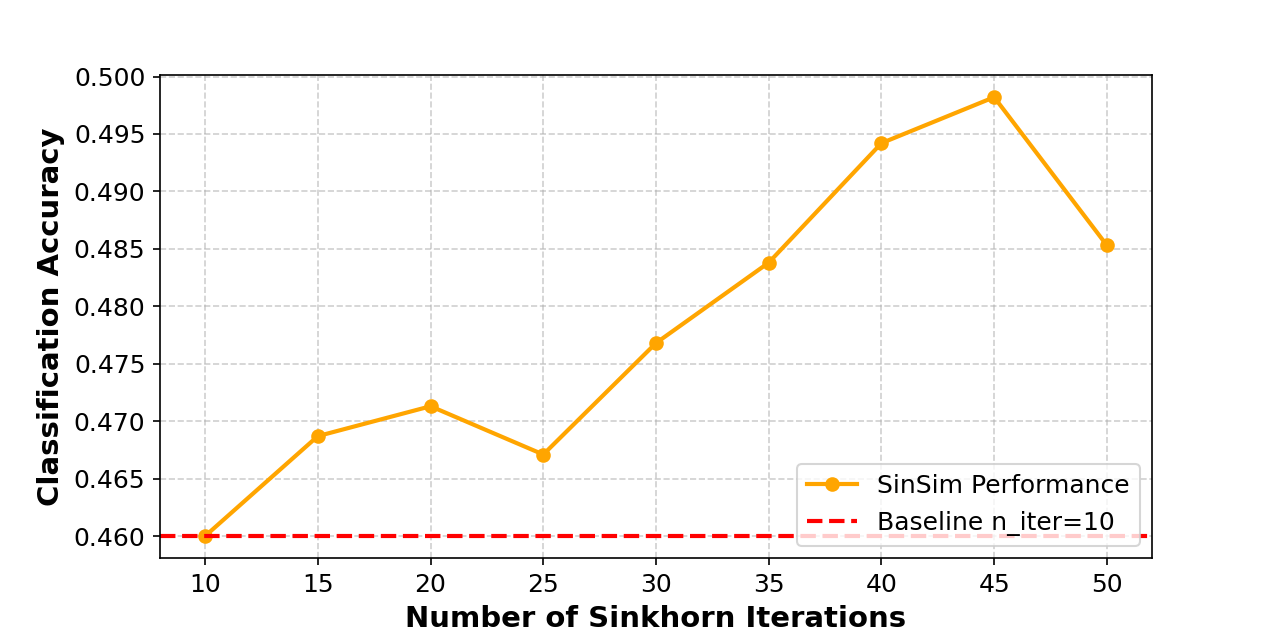}
    \caption{Effect of Sinkhorn iterations, as in Figure~\ref{fig:n_iter}, but now assessed on CIFAR-10. The overall trend is very similar to that of MNIST, with slightly increased variability, but consistent overall conclusion.}
    \label{fig:n_iter_cifar}
\end{figure}

Figure~\ref{fig:n_iter} illustrates the classification accuracy across different values of iteration for MNIST, while Figure~\ref{fig:n_iter_cifar} shows the same effect on CIFAR-10. The results show that increasing the number of iterations improves performance up to an optimal value (around 40), after which performance slightly degrades. This trend suggests that while Sinkhorn optimization is beneficial, excessive iterations may lead to over-smoothing, potentially reducing the discriminative power of learned representations. 

\subsubsection*{Effect of Regularization Strength $\lambda$}

The hyperparameter $\lambda$ in the Sinkhorn regularization term controls the trade-off between entropy regularization and transport cost. A low $\lambda$ results in minimal regularization, while a high $\lambda$ may overly smooth the feature space, leading to a loss of discriminative power.

\begin{figure}[h]
    \centering
    \includegraphics[width=1\linewidth]{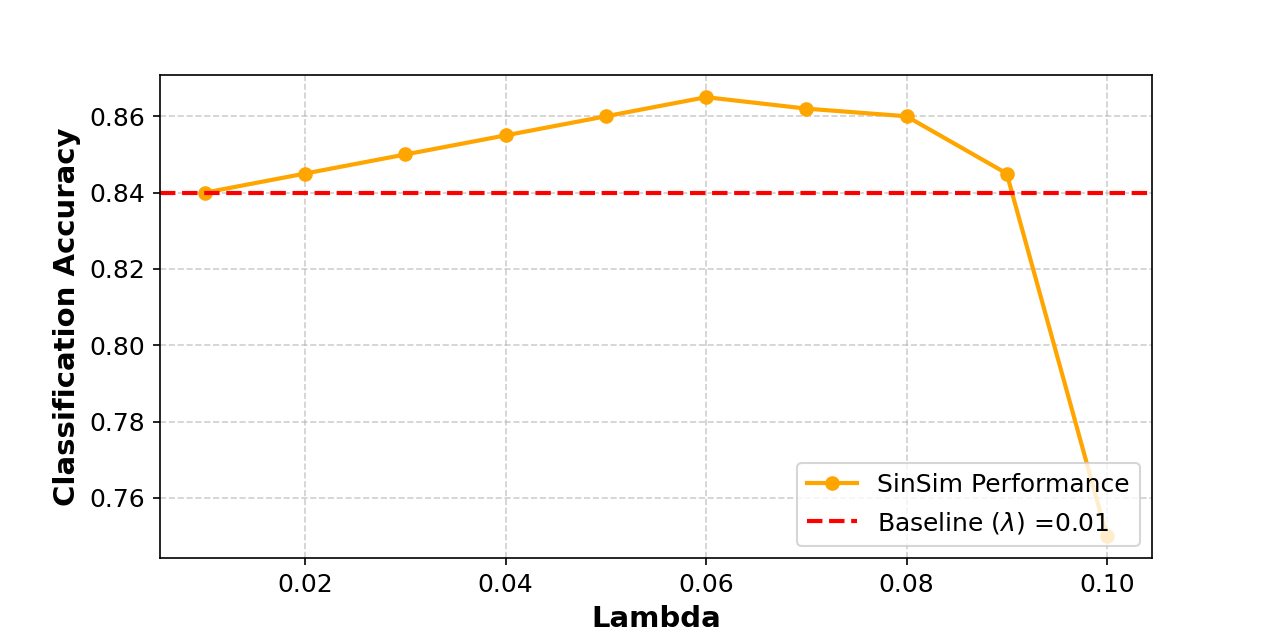}
    \caption{Effect of Sinkhorn regularization strength $\lambda$ on SinSim classification accuracy on MNIST. Moderate values ($\lambda=0.06$) yield the best performance, while overly large values reduce discriminative power. The red dashed line represents a default baseline at $\lambda=0.01$.}
    \label{fig:lambda}
\end{figure}

\begin{figure}[h]
    \centering
    \includegraphics[width=1\linewidth]{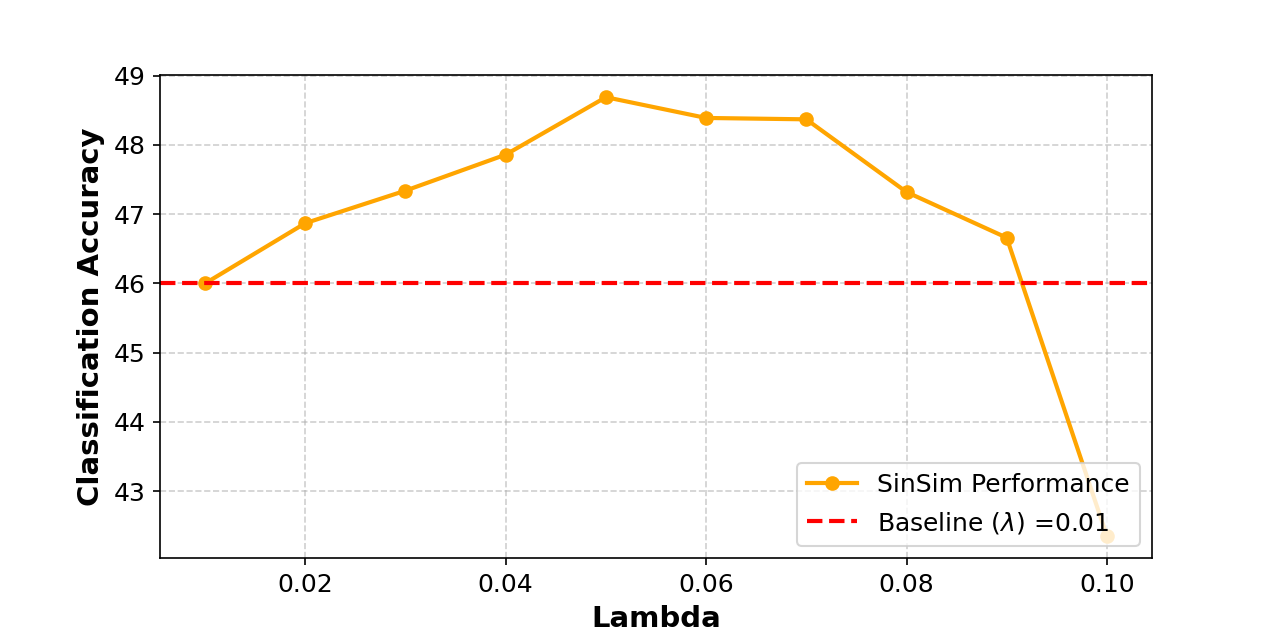}
    \caption{Effect of Sinkhorn regularization strength $\lambda$, as in Figure~\ref{fig:lambda}, but here tested on CIFAR-10, with results and overall conclusion similar to those found for MNIST.}
    \label{fig:lambda_cifar}
\end{figure}

Figure~\ref{fig:lambda} presents the classification accuracy as a function of $\lambda$ for MNIST, and Figure~\ref{fig:lambda_cifar} for CIFAR-10. The results for both datasets give consistent results, with a moderate value of $\lambda$ (around $\lambda = 0.06$) achieving the best performance. For very small values, the impact of Sinkhorn regularization is too limited, while for large values ($\lambda > 0.09$), performance sharply drops, confirming that excessive smoothing negatively affects representation quality. 

\section{Conclusion}

We have introduced SinSim, a novel self-supervised learning framework that integrates Sinkhorn regularization into the SimCLR paradigm. By leveraging optimal transport theory, SinSim enforces a structured latent space while preserving the discriminative power of contrastive learning. Our results demonstrate that SinSim not only enhances representation structure but also improves classification performance, particularly in settings where contrastive learning alone struggles to maintain global alignment.

Through both theoretical insights and empirical evaluations, we have shown that Sinkhorn-based contrastive learning achieves superior representation quality, as measured by both linear and non-linear evaluation protocols. Moreover, our UMAP visualizations reveal that Sinkhorn regularization promotes a well-dispersed and semantically meaningful feature space, reducing representation collapse and improving class separation.

Future work will explore the extension of Sinkhorn regularization to other self-supervised methods and investigate its impact on larger-scale datasets and more complex architectures. A promising direction is examining its effectiveness in multimodal learning settings, such as vision-language models, where structured feature alignment is critical. Additionally, the structured feature representations encouraged by SinSim can benefit real-world applications such as medical imaging and robotics, where robust and geometry-aware representations are essential.

\end{document}